\documentclass[11pt]{article}
\usepackage[margin=1in]{geometry}
\usepackage{amsmath,amssymb,amsthm}
\usepackage{newtxtext}
\usepackage[subscriptcorrection]{newtxmath}
\usepackage[round,authoryear]{natbib}
\usepackage{hyperref}
\hypersetup{hidelinks}
\newcommand{\cX}{\mathcal{X}}
\newcommand{\cY}{\mathcal{Y}}
\newcommand{\cD}{\mathcal{D}}
\newcommand{\cA}{\mathcal{A}}
\newcommand{\fX}{\mathfrak{X}}
\newcommand{\Prob}{\mathbb{P}}
\newcommand{\E}{\mathbb{E}}
\newcommand{\ind}{\mathbf{1}}
\newcommand{\VC}{\mathrm{VC}}

\newtheorem{theorem}{Theorem}

\newtheorem{corollary}{Corollary}
\theoremstyle{definition}
\newtheorem{definition}{Definition}
\theoremstyle{remark}
\newtheorem{remark}{Remark}

\newenvironment{keywords}{\par\noindent\textbf{Keywords: }}{\par}

\title{A note on conditional PAC-efficient reasoning in large language model routing}
\author{
Hao Zeng\thanks{Department of Statistics and Data Science, Southern University of Science and Technology, Shenzhen, China. Email: \texttt{zengh@sustech.edu.cn}.}
\and
Bingyi Jing\thanks{School of Artificial Intelligence, The Chinese University of Hong Kong, Shenzhen, China, and Shenzhen Loop Area Institute, Shenzhen, China. Email: \texttt{bingyijing@cuhk.edu.cn}.}
}
\date{}

\begin{document}

\maketitle

\begin{abstract}
We study distribution-free risk control for model routing, motivated by large language model reasoning.
We formalize pointwise conditional efficiency under a probably approximately correct guarantee and show that it forces a nearly impossible router: at almost every input where the fast model exceeds the target loss, the algorithm must route to the expert with probability at least one minus the prescribed error level.
We therefore introduce a restricted conditional formulation based on a prespecified family of conditioning sets, together with an explicit router.
The proposed router achieves finite-sample conditional validity and, under separation and margin conditions, near-oracle expert usage.
The main insight is that the level of conditioning determines whether distribution-free reliability can coexist with computational savings: pointwise control is too strong, whereas structured setwise control remains feasible.
\end{abstract}


\section{Introduction}

Large language models (LLMs) are increasingly studied in statistics, both as tools for data analysis and as systems whose alignment, watermarking, and reliability raise statistical issues~\citep{sun2025lambda,xiao2025algorithmic,xie2025debiasing,cai2026better}.
Their practical appeal is clear: recent LLMs such as DeepSeek-R1, Qwen3, GPT-5, and Llama~3 perform well on complex problem-solving tasks~\citep{deepseek-ai2025deepseekr1,yang2025qwen3,openai2025introducing,grattafiori2024llama}, but they remain expensive at deployment time~\citep{kwon2023efficient}.
A common source of this cost is overthinking, where a reasoning model spends long chains of thought (CoT) on easy inputs that could be handled by a cheaper or shorter reasoning mode~\citep{sui2025stop,aggarwal2025optimalthinkingbench}.
This motivates work on efficient reasoning, including model routing~\citep{ong2025routellm,dekoninck2025unified}, speculative decoding~\citep{leviathan2023fast}, and adaptive reasoning strategies~\citep{snell2024scaling}.
These methods can reduce computational cost, but they usually do not provide finite-sample statistical control of the resulting performance loss.

To address this gap, {\bf PAC reasoning}~\citep{zeng2026provable} studies routing between an expensive expert model and a cheaper fast model under a distribution-free guarantee on the performance loss.
The marginal guarantee controls unsafe fast-model use only on average, allowing larger error rates on particular inputs or subgroups.
For example, a router may be safe on easy questions while sending a hard subgroup to the fast model with uncontrolled loss.
We ask whether stronger conditional PAC-style control can coexist with a non-trivial reduction in expert-model usage.
Recent work has studied related PAC-style guarantees for reasoning and human-LLM systems~\citep{huang2026conditional,yu2026anytime,zeng2026hypac}.
The same tension is familiar from distribution-free conformal prediction: exact conditional guarantees are impossible without additional assumptions, whereas restricted, local, selective, or groupwise guarantees can still be feasible~\citep{vovk2012conditional,lei2014distributionfree,lei2018distributionfree,barber2021limits,guan2023localized,gibbs2025conformal,hore2025conformal,bao2024selective,barber2023conformal}.
Related finite-sample tools also show how conformal ideas can support conditional or simultaneous error control in testing, computation, and model selection~\citep{bates2023testing,hu2023twosample,lei2019fast,yang2024selection,liang2025algorithmica}.

In this work, we first formalize pointwise conditional PAC efficiency and characterize its cost in the distribution-free setting.
The result shows that any pointwise valid router must defer to the expert with high probability at almost every input where the fast model exceeds the target loss.
This gives a sharp boundary for distribution-free PAC routing: on inputs where the fast model exceeds the target loss, it can be used with probability at most $\alpha$.
We then introduce a {\bf restricted conditional} formulation indexed by a prespecified family of conditioning sets.
For this relaxed target, we construct an explicit router and establish both finite-sample validity and a near-oracle expert-usage bound.
The formulation also covers groupwise and fixed-radius local guarantees as special cases.
Together, these results identify a theoretical boundary for LLM reasoning acceleration with statistical guarantees and provide a feasible setwise relaxation.
They also offer a broader lesson for predictive inference: the choice of conditioning level determines the trade-off between distribution-free validity and practical efficiency.
All proofs and supporting technical results are provided in the Supplementary Material.

\section{PAC routing}
\label{sec:setup}

Let $\cX$ be the input space and $\cY$ the output space.
We observe data from a distribution $P$ on $\cX \times \cY$, with $P_X$ denoting the marginal law of $X$.
Given $x \in \cX$, an expert model $f : \cX \to \cY$, such as DeepSeek-R1, produces $f(x)$.
A fast model $\tilde f : \cX \to \cY$, such as DeepSeek-V3, produces $\tilde f(x)$.
We measure loss through a function $\ell : \cY \times \cY \to [0,\infty)$.
We use a held-out calibration sample $\cD_{\mathrm{cal}}=\{X_i\}_{i=1}^n\sim P_X^n$.
Let $g:\cX \to \{0,1\}$ be a routing rule, where $g(x)=1$ sends the input to the expert model and $g(x)=0$ uses the fast model.
To reduce inference cost or latency, we combine the two models through the routing rule into the composite predictor
\[
\hat f(x)=
\begin{cases}
f(x), & \text{if } g(x)=1,\\
\tilde f(x), & \text{if } g(x)=0.
\end{cases}
\]

Fix a target loss level $\epsilon>0$ and an error level $\alpha\in(0,1)$.
Define
\[
L(x)=\ind\{\ell(\tilde f(x),f(x))>\epsilon\}.
\]
Thus $L(x)=1$ means that the fast model exceeds the target loss level at input $x$, and $L(x)=0$ means that it stays within the target.

\begin{remark}
The loss function $\ell$ can be chosen according to the task.
For open-ended answers, $\ell$ may measure semantic dissimilarity between the fast-model answer and the expert-model answer, for example by one minus an embedding similarity score.
For classification or multiple-choice tasks, $\ell$ can be the 0--1 loss that records whether the two models give different labels or options.
For graded reasoning tasks, $\ell$ can also be a bounded evaluation loss produced by a verifier or a reward model.
\end{remark}

Since an error matters only when the fast model is used, the relevant pointwise risk of a routing rule is
\[
R(\hat f;x)=\ind\{g(x)=0\}\,L(x).
\]
For a fixed router, its expert usage probability is $\Prob_{X\sim P_X}(g(X)=1)$, equivalently $\E_{X\sim P_X}[g(X)]$.

\begin{definition}[Marginal PAC efficiency]
\label{def:marginal}
An algorithm $\cA$ is $(\epsilon,\alpha)$-marginally PAC efficient if for all distributions $P$,
\[
\Prob_{\cD_{\mathrm{cal}} \sim P_X^n,\ X \sim P_X}\bigl(R(\hat f;X)=1\bigr)\le \alpha.
\]
\end{definition}

This is the marginal guarantee studied in PAC reasoning~\citep{zeng2026provable}.
The probability in Definition~\ref{def:marginal} is over both the calibration sample and a test sample.
Thus, the guarantee controls the population-average failure rate of the deployed router.
The next definition asks for the same control at almost every input point.

\begin{definition}[Pointwise conditional PAC (CPAC) efficiency]
\label{def:pointwise_conditional}
An algorithm $\cA$ is $(\epsilon,\alpha)$-pointwise conditionally PAC efficient if for all distributions $P$ and for $P_X$-almost every $x\in\cX$,
\[
\Prob_{\cD_{\mathrm{cal}} \sim P_X^n}\bigl(R(\hat f;x)=1\bigr)\le \alpha.
\]
\end{definition}

In Definition~\ref{def:pointwise_conditional}, the input $x$ is fixed before the calibration sample is drawn.
The probability is therefore only over the calibration randomness used to build the router.
This is much stronger than marginal PAC efficiency because the algorithm must protect almost every input separately, not only on average.

\section{Impossibility of pointwise conditional PAC efficiency}
\label{sec:impossibility}

We first show that Definition~\ref{def:pointwise_conditional} is too strong in the distribution-free setting.

\begin{theorem}[Impossibility of pointwise conditional PAC efficiency]
\label{thm:impossibility}
Let
\[
E_\epsilon=\{x\in\cX:L(x)=1\}=\{x\in\cX:\ell(\tilde f(x),f(x))>\epsilon\}.
\]
Then an algorithm $\cA$ is $(\epsilon,\alpha)$-pointwise conditionally PAC efficient if and only if for all distributions $P$ and for $P_X$-almost every $x\in E_\epsilon$,
\[
\Prob_{\cD_{\mathrm{cal}} \sim P_X^n}\bigl(g(x)=0\bigr)\le \alpha.
\]
\end{theorem}

\begin{remark}
Theorem~\ref{thm:impossibility} implies that a pointwise conditional guarantee forces the router to use the fast model with probability at most $\alpha$ at almost every input where the fast model exceeds the target loss.
In other words, exact conditional PAC efficiency allows \textit{no acceleration}: the fast model can be used with probability at most $\alpha$ on almost every input where it exceeds the target loss.
It motivates us to consider weakening the conditioning event so that it ranges only over a user-specified restricted family of sets.
\end{remark}

\section{Restricted conditional PAC routing}
\label{sec:restricted}

The impossibility characterization above motivates feasible conditional-style guarantees.
We therefore replace single-input conditions with a prespecified family $\fX$ of sets.
Pointwise conditional PAC efficiency controls almost every input, whereas restricted conditional PAC efficiency controls sets in $\fX$ that have sufficient probability mass.
This keeps a conditional interpretation while relaxing the pointwise restriction in Theorem~\ref{thm:impossibility}.

We next give a concrete construction for this restricted target.
For simplicity of analysis, we focus below on \textit{threshold-based routing rules}.
The impossibility result in Section~\ref{sec:impossibility} applies to general routing rules, while the construction uses uncertainty-threshold routers as a simple and analyzable subclass.
Threshold-based routing is also the basic form used in several PAC reasoning and related conformal routing methods, where a score is compared with a calibrated cutoff to decide whether the expert model is needed~\citep{zeng2026provable,huang2026conditional,yu2026anytime,zeng2026hypac,su2025cprouter}.
Let $s : \cX \to \mathbb{R}$ be an uncertainty score, treated as fixed before calibration.
Intuitively, the score should rank inputs by the loss risk of the fast model: larger values should indicate that $\ell(\tilde f(x),f(x))$ is more likely to exceed the target level.
Examples include logits-based, verbalized, router-based, and sampling-based uncertainty scores; see the Supplementary Material for concrete constructions.
The validity result only requires this score to be fixed before calibration.
The quality of the score mainly affects power, or equivalently expert usage, because a score that better tracks the loss function can defer fewer safe inputs while still controlling the restricted conditional risk.

\begin{definition}[Restricted conditional PAC (RCPAC) efficiency]
\label{def:restricted}
Fix $\delta,\gamma \in (0,1)$ and a family $\fX$ of measurable subsets of $\cX$.
Assume that all displayed suprema, infima, and induced routing rules are measurable.
For a fixed calibration sample, write $\hat f_{\cD_{\mathrm{cal}}}$ for the resulting composite predictor and, for any $A\in\fX$ with $P_X(A)>0$, define
\[
q_A(\cD_{\mathrm{cal}})
=
\Prob_{X\sim P_X}\bigl(R(\hat f_{\cD_{\mathrm{cal}}};X)=1\mid X\in A\bigr).
\]
An algorithm $\cA$ is $(\epsilon,\alpha,\delta,\gamma,\fX)$-restricted conditional PAC efficient if for every distribution $P$,
\[
\Prob_{\cD_{\mathrm{cal}} \sim P_X^n}\Bigl(
\sup_{A\in\fX:\,P_X(A)\ge \delta}
q_A(\cD_{\mathrm{cal}})>\alpha
\Bigr)\le \gamma.
\]
The supremum over an empty admissible family is defined as $-\infty$.
\end{definition}

\begin{remark}
Definition~\ref{def:restricted} replaces conditioning on individual inputs with conditioning on prespecified sets $A\in\fX$ with $P_X(A)\ge\delta$.
However, RCPAC and pointwise CPAC are not formally ordered because their calibration-sample and test-input quantifiers differ.
Thus RCPAC does not in general reduce to pointwise CPAC, even when the admissible sets become small.
At the other extreme, if $\{A\in\fX:P_X(A)\ge\delta\}=\{\cX\}$, then RCPAC becomes
\[
\Prob_{\cD_{\mathrm{cal}}\sim P_X^n}\bigl(q_{\cX}(\cD_{\mathrm{cal}})>\alpha\bigr)\le\gamma
\quad\Longrightarrow\quad
\E_{\cD_{\mathrm{cal}}\sim P_X^n}\bigl[q_{\cX}(\cD_{\mathrm{cal}})\bigr]
\le \alpha+(1-\alpha)\gamma.
\]
At the boundary $\gamma=0$, if this value is allowed, RCPAC implies the marginal PAC target but remains stronger than Definition~\ref{def:marginal}.
Therefore RCPAC does not exactly reduce to marginal PAC efficiency.
\end{remark}

We now construct the {\bf Restricted Conditional PAC Router (RC-PAC Router)}.
Fix $(\epsilon,\alpha,\delta,\gamma,\fX)$, an uncertainty score $s$, and a calibration slack $\eta_n>0$.
Let $\overline{\mathbb{R}}=\mathbb{R}\cup\{-\infty,+\infty\}$.
For any measurable threshold function $t:\cX\to\overline{\mathbb{R}}$, define
\[
g_t(x)=\ind\{s(x)\ge t(x)\},
\]
where a scalar threshold is identified with a constant function and $g_{-\infty}(x)=1$, $g_{+\infty}(x)=0$.
For $A\in\fX$ with $P_X(A)>0$ and $\tau\in\mathbb{R}$, define the population restricted risk
\[
r_P(A,\tau)=\Prob_{X\sim P_X}\bigl(L(X)=1,\ s(X)<\tau\mid X\in A\bigr).
\]
Extend $r_P(A,\tau)$ to $\tau\in\overline{\mathbb{R}}$ by monotone limits.
For each $A\in\fX$, let
\begin{equation}
N(A)=\sum_{i=1}^n \ind\{X_i \in A\}.
\label{eq:set-count}
\end{equation}
For every set $A\in\fX$ and threshold $\tau\in\mathbb{R}$, define the empirical restricted conditional risk
\[
\hat r(A,\tau)=\left(\sum_{i=1}^n \ind\{X_i\in A,\ s(X_i)<\tau,\ L(X_i)=1\}+\ind\{N(A)=0\} \right) /\max\{N(A),1\}.
\]
and the calibrated setwise threshold in $\overline{\mathbb{R}}$
\begin{equation}
\hat\tau(A)=\sup\bigl\{\tau\in\mathbb{R} : \hat r(A,\tau)+\eta_n \le \alpha \bigr\}.
\label{eq:setwise-threshold}
\end{equation}
We use the convention $\sup\emptyset=-\infty$.
For a test input $x\in\cX$, aggregate these setwise thresholds through
\begin{equation}
\hat\tau(x)=\inf\bigl\{\hat\tau(A) : A\in\fX,\ x\in A \bigr\},
\label{eq:adaptive-threshold}
\end{equation}
with the convention $\hat\tau(x)=-\infty$ if the set on the right-hand side is empty.
The RC-PAC Router then defers to the expert when $s(x)\ge\hat\tau(x)$, that is,
\begin{equation}
g_{\hat\tau}(x)=\ind\{s(x)\ge\hat\tau(x)\}.
\label{eq:rcpac-router}
\end{equation}

For technical convenience in the validity proof, define
\[
\fX_s = \bigl\{ A \cap \{x\in\cX : s(x)< t\} : A\in\fX,\ t\in\mathbb{R} \bigr\}.
\]
Its complexity is summarized by the VC dimension $d_s=\VC(\fX_s)$; see the Supplementary Material for the formal definition.
The proof of the restricted validity theorem uses a relative VC concentration bound and three routing lemmas stated formally in the Supplementary Material.
The routing lemmas give a uniform calibration event over the induced class $\fX_s$, setwise validity of the calibrated thresholds, and a monotonicity argument for the inf-envelope aggregation.
Combining these ingredients yields the main finite-sample guarantee below.

\begin{theorem}[Restricted conditional validity of the RC-PAC Router]
\label{thm:restricted_validity}
Assume that the score-induced class has finite VC dimension, and set
$
d_s=\VC(\fX_s),
\kappa_n=d_s\log(en)+\log(4/\gamma).
$
There exist universal constants $c_{\mathrm{vc}},C_{\mathrm{vc}}>0$ such that, if
$
n\delta\ge c_{\mathrm{vc}}\kappa_n,
\eta_n=C_{\mathrm{vc}}\sqrt{\frac{\kappa_n}{n\delta}},
$
then the router defined by \eqref{eq:set-count}--\eqref{eq:rcpac-router} is $(\epsilon,\alpha,\delta,\gamma,\fX)$-restricted conditional PAC efficient.
Moreover, with probability at least $1-\gamma$ over $\cD_{\mathrm{cal}}\sim P_X^n$, every $A\in\fX$ with $P_X(A)\ge\delta$ satisfies
\[
N(A)\ge n\delta/2,
\qquad
\Prob_{X\sim P_X}\bigl(L(X)=1,\ g_{\hat\tau}(X)=0\mid X\in A\bigr)\le\alpha.
\]
\end{theorem}

Theorem~\ref{thm:restricted_validity} is a validity result.
To compare efficiency, we next define oracle thresholds.
Define the oracle setwise threshold in $\overline{\mathbb{R}}$
\begin{equation}
\tau^\star(A)=\sup\{\tau \in \mathbb{R} : r_P(A,\tau)\le \alpha\}.
\label{eq:oracle-set-threshold}
\end{equation}
Define the population-admissible calibrated local threshold as follows:
\begin{equation}
\hat\tau_\delta(x)=\inf\bigl\{\hat\tau(A): A\in\fX,\ x\in A,\ P_X(A)\ge \delta\bigr\}
\label{eq:population-admissible-threshold}
\end{equation}
Define the oracle local threshold as follows:
\begin{equation}
\tau^\star(x)=\inf\bigl\{\tau^\star(A): A\in\fX,\ x\in A,\ P_X(A)\ge \delta\bigr\}
\label{eq:oracle-local-threshold}
\end{equation}
Define $\fX_\delta(x)$ and $\cX_\delta$ as follows:
\[
\fX_\delta(x)=\{A\in\fX:x\in A,\ P_X(A)\ge\delta\},
\qquad
\cX_\delta=\{x\in\cX:\fX_\delta(x)\ne\emptyset\}.
\]
Here ``local'' refers to the threshold aggregated at the point $x$, whereas ``restricted'' refers to the construction induced by the family $\fX$.
The threshold $\hat\tau_\delta$ defined in \eqref{eq:population-admissible-threshold} is used only for oracle comparison, while the implemented RC-PAC Router uses the data-adaptive threshold $\hat\tau$ in \eqref{eq:adaptive-threshold}.
We use the convention $\hat\tau_\delta(x)=\tau^\star(x)=-\infty$ if the population-admissible set family is empty.
The oracle restricted router is $g_{\tau^\star}(x)=\ind\{s(x)\ge\tau^\star(x)\}$.
For a fixed calibration sample, $\E_X$ denotes expectation over an independent $X\sim P_X$.
Let $\mathcal E_n$ denote the uniform calibration event
\[
\mathcal E_n
=
\left\{N(A)\ge n\delta/2,\ \sup_{\tau\in\mathbb{R}}\bigl|\hat r(A,\tau)-r_P(A,\tau)\bigr|\le\eta_n,\ \forall A\in\fX\text{ with }P_X(A)\ge\delta\right\}.
\]
The proof of Theorem~\ref{thm:restricted_validity} shows that $\Prob_{\cD_{\mathrm{cal}}\sim P_X^n}(\mathcal E_n)\ge1-\gamma$.
Define the high-level small-set nonactivity event
\[
\mathcal S_n
=
\bigl\{\hat\tau(x)=\hat\tau_\delta(x)
\text{ for }P_X\text{-almost every }x\bigr\}.
\]
The event $\mathcal S_n$ says that sets below the mass threshold do not lower the implemented local threshold.
It is an additional algorithmic condition and does not follow from the VC bound alone.
We next relate the calibrated thresholds to the oracle thresholds in \eqref{eq:oracle-set-threshold} and \eqref{eq:oracle-local-threshold}.
For this step, a pure monotonicity assumption is not enough.
We also need a one-sided separation condition, which means that moving the threshold below $\tau^\star(A)$ keeps the risk below $\alpha$ by a margin.

\begin{theorem}[Near-oracle expert usage]
\label{thm:oracle_efficiency}
Assume the conditions of Theorem~\ref{thm:restricted_validity} and suppose that $\Prob_{\cD_{\mathrm{cal}}\sim P_X^n}(\mathcal S_n)\ge1-\rho_n$ for some $\rho_n\in[0,1-\gamma)$.
For constants $\lambda,C_0,\beta,t_0>0$, assume that every $A\in\fX$ with $P_X(A)\ge\delta$ has $\tau^\star(A)\in\mathbb{R}$ and satisfies
\[
r_P\bigl(A,\tau^\star(A)-t\bigr)\le\alpha-\lambda t,
\qquad t\in[0,t_0].
\]
Assume that $\tau^\star(x)\in\mathbb{R}$ for $P_X$-almost every $x\in\cX_\delta$ and
\[
P_X\bigl\{x\in\cX_\delta:|s(x)-\tau^\star(x)|\le u\bigr\}
\le C_0u^\beta,
\qquad u\in[0,t_0].
\]
If $2\eta_n\le\min\{\alpha,\lambda t_0\}$, then, with probability at least $1-\gamma-\rho_n$ over the calibration sample,
\[
0\le\tau^\star(x)-\hat\tau(x)\le\frac{2\eta_n}{\lambda}
\]
for $P_X$-almost every $x\in\cX_\delta$, and
\[
\E_X[g_{\hat\tau}(X)]
\le
\E_X[g_{\tau^\star}(X)]
+C_0\bigl(2\eta_n/\lambda\bigr)^\beta.
\]
On $\cX_\delta^c$, both routers defer on $\mathcal S_n$.
\end{theorem}

\begin{remark}
Theorem~\ref{thm:oracle_efficiency} does not claim that the RC-PAC Router is globally optimal.
Instead, it shows that once restricted conditional validity is feasible, the price to buy finite-sample calibration is small when the oracle threshold is separated and the uncertainty-score distribution has little mass near the oracle decision boundary.
\end{remark}

The next two subsections illustrate this framework in two important cases: group-conditional routing and fixed-radius local routing.

\subsection{Group-conditional special case}
Group-conditional PAC routing is obtained by taking $\fX$ to be a finite partition.
This makes it a direct corollary of the restricted conditional theory above.
Related finite-partition or classwise guarantees also appear in conformal set-valued classification with confidence control~\citep{lei2014classification,sadinle2019least}.

\begin{corollary}[Group-conditional PAC routing]
\label{cor:group}
Let $h:\cX \to \{1,\dots,K\}$ be a taxonomy, and let
\[
\fX=\{G_1,\dots,G_K\}, \qquad G_k=\{x\in\cX:h(x)=k\}.
\]
Assume the conditions of Theorem~\ref{thm:restricted_validity} hold for this family and run the RC-PAC Router with the corresponding $\eta_n$.
Then, with probability at least $1-\gamma$ over $\cD_{\mathrm{cal}}\sim P_X^n$,
\[
\Prob_{X\sim P_X}\bigl(L(X)=1,\ g_{\hat\tau}(X)=0 \mid h(X)=k\bigr)\le \alpha
\]
for every group $k$ such that $P_X(G_k)\ge \delta$.
\end{corollary}

\begin{remark}
For a finite taxonomy, each of the $K$ groups induces at most $m+1$ threshold traces on any $m$ points, so $\Pi_{\fX_s}(m)\le K(m+1)$.
Thus the same relative VC argument replaces the generic VC complexity term by one of order $\log K+\log n$.
The main message is structural: group-conditional PAC efficiency is not a separate theory, but the simplest restricted conditional special case, including the group-level conditional PAC setting~\citep{huang2026conditional}.
\end{remark}

\subsection{Local guarantees and limits}
Let $d$ be a metric on $\cX$.
Fix $r>0$ and assume that every closed ball $B(x,r)=\{z\in\cX:d(z,x)\le r\}$ is measurable.
Let
\[
\fX_r=\{B(x,r):x\in\cX\}.
\]

\begin{corollary}[Fixed-radius local guarantees]
\label{cor:local_as_restricted}
If the router defined by \eqref{eq:set-count}--\eqref{eq:rcpac-router} is $(\epsilon,\alpha,\delta,\gamma,\fX_r)$-restricted conditional PAC efficient, then, with probability at least $1-\gamma$ over $\cD_{\mathrm{cal}}\sim P_X^n$,
\[
\Prob_{X\sim P_X}\bigl(L(X)=1,\ g_{\hat\tau}(X)=0 \mid X\in B(x,r)\bigr)\le \alpha
\]
for every $x\in\cX$ such that $P_X(B(x,r))\ge \delta$.
\end{corollary}

\begin{remark}
At a fixed radius $r$, local PAC routing is therefore a special case of restricted conditional PAC routing.
The situation changes when the neighborhood scale varies with $n$ or depends on the calibration sample.
If $r_n \to 0$, the family becomes $\fX_{r_n}=\{B(x,r_n):x\in\cX\}$, and both the mass threshold and the complexity term may vary with $n$.
For $k$-nearest-neighbor neighborhoods, the family $\fX^{\mathrm{kNN}}_{n,k}=\{\mathcal N_{n,k}(x):x\in\cX\}$ is random, where $\mathcal N_{n,k}(x)$ denotes the set of the $k$ nearest calibration points to $x$ under a fixed metric on $\cX$.
To obtain shrinking-radius or data-dependent local guarantees, one needs additional control of class complexity, local mass, and approximation bias, as in local conformal inference~\citep{lei2014distributionfree,guan2023localized,hore2025conformal,candes2023conformalized,gui2024conformalized,lunde2025conformal}.
An important open question is whether restricted conditional PAC guarantees can be established for richer routing classes beyond uncertainty-threshold rules.
\end{remark}

\section{Conclusion}

We study how distribution-free validity constrains inference efficiency in PAC reasoning.
Our impossibility result shows that pointwise conditional validity forces a nearly trivial router: at almost every input where the fast model exceeds the target loss, the expert must be used with probability at least $1-\alpha$ over the calibration sample.
To obtain a useful guarantee, we replace pointwise conditioning with conditioning on prespecified sets of sufficient population mass.
For families with bounded score-induced complexity and sufficient effective sample size, the resulting RC-PAC Router achieves finite-sample validity and near-oracle expert usage, with group-conditional and fixed-radius local guarantees as special cases.
The central insight is that the level of conditioning, rather than calibration alone, determines whether distribution-free reliability can coexist with computational savings.
Pointwise guarantees are too strong to permit useful acceleration without further assumptions, whereas restricted guarantees recover efficiency by protecting sets that are large and structured enough to support reliable calibration.
This principle extends beyond large language model routing to the broader design of distribution-free predictive inference.

\section*{Supplementary Material}

The Supplementary Material includes concrete constructions of uncertainty scores and all proofs.

\nocite{sauer1972density}
\nocite{vapnik1971uniform}
\nocite{hoeffding1963probability}
\bibliography{note_pac}

\end{document}